\documentclass[]{icas2026} 

\usepackage[english]{babel}
\usepackage{amsmath}
\usepackage{amssymb}
\usepackage{amsfonts}
\usepackage{algorithmic}
\usepackage{graphicx}
\usepackage{svg}
\usepackage{float}
\usepackage{textcomp}
\usepackage[dvipsnames]{xcolor}
\usepackage{soul}
\usepackage{csquotes}

\usepackage{microtype}

\usepackage{tabularx}
\usepackage{booktabs}

\usepackage{placeins}

\usepackage{hyperref}
\usepackage[capitalize,noabbrev]{cleveref}

\usepackage{siunitx}
\DeclareSIUnit\knot{kn}
\DeclareSIUnit\nauticalmile{NM}
\DeclareSIUnit\foot{ft}
\DeclareSIUnit\flops{FLOPS}

\usepackage{bbding}
\usepackage{tikz}
\usepackage[edges]{forest}
\usepackage{wheelchart}
\usepackage{tikz}
\usetikzlibrary{arrows.meta, positioning, shapes.geometric}

\tikzset{
block/.style = {rectangle, draw, rounded corners,
                minimum height=0.9cm, minimum width=3cm,
                align=center},
decision/.style = {diamond, draw, aspect=2,
                   align=center, inner sep=1pt},
data/.style = {trapezium, draw,
               trapezium left angle=70,
               trapezium right angle=110,
               align=center},
line/.style = {draw, -{Latex[length=3mm]}, thick}
}

\usepackage{pgfplots}
\usepackage{pgfplotstable}
\usepgfplotslibrary{groupplots}
\pgfplotsset{compat=newest}
\usepgfplotslibrary{
    fillbetween,
    statistics
}

\definecolor{DLRBlack}{gray}{0}
\definecolor{DLRGrey}{gray}{0.420} 
\colorlet{DLRGray}{DLRGrey}
\definecolor{DLRWhite}{gray}{1}

\colorlet{DLREagleColor}{DLRBlack}
\colorlet{DLRTextColor}{DLRBlack}

\colorlet{DLRDarkerGrey}{DLRGrey}
\definecolor{DLRDarkGrey}{gray}{0.537} 
\definecolor{DLRMediumGrey}{gray}{0.702} 
\definecolor{DLRLightGrey}{gray}{0.820} 
\definecolor{DLRLighterGrey}{gray}{0.929} 

\colorlet{DLRDarkerGray}{DLRDarkerGrey}
\colorlet{DLRDarkGray}{DLRDarkGrey}
\colorlet{DLRMediumGray}{DLRMediumGrey}
\colorlet{DLRLightGray}{DLRLightGrey}
\colorlet{DLRLighterGray}{DLRLighterGrey}

\definecolor{DLRDarkerBlue}{RGB}{0, 106, 144}
\definecolor{DLRDarkBlue}{RGB}{0, 156, 208}
\definecolor{DLRBlue}{RGB}{33, 187, 223}
\colorlet{DLRMediumBlue}{DLRBlue}
\definecolor{DLRLightBlue}{RGB}{149, 212, 238}
\definecolor{DLRLighterBlue}{RGB}{201, 232, 251}

\definecolor{DLRDarkerGreen}{RGB}{115, 163, 63}
\definecolor{DLRDarkGreen}{RGB}{158, 193, 76}
\definecolor{DLRGreen}{RGB}{199, 214, 84}
\colorlet{DLRMediumGreen}{DLRGreen}
\definecolor{DLRLightGreen}{RGB}{215, 223, 116}
\definecolor{DLRLighterGreen}{RGB}{228, 234, 173}

\definecolor{DLRDarkerYellow}{RGB}{224, 177, 57}
\definecolor{DLRDarkYellow}{RGB}{254, 206, 73}
\definecolor{DLRYellow}{RGB}{255, 223, 73}
\colorlet{DLRMediumYellow}{DLRYellow}
\definecolor{DLRLightYellow}{RGB}{255, 234, 117}
\definecolor{DLRLighterYellow}{RGB}{255, 248, 189}

\definecolor{DLRDarkestBlue}{RGB}{0, 50, 69}
\definecolor{DLRDarkestGreen}{RGB}{99, 119, 34}
\definecolor{DLRDarkestYellow}{RGB}{190, 150, 0}

\definecolor{DLRRed}{RGB}{179, 63, 61}
\colorlet{DLRMediumRed}{DLRRed}
\definecolor{DLRLightRed}{RGB}{199, 122, 109}
\definecolor{DLRLighterRed}{RGB}{223, 180, 168}

\definecolor{DLRDarkestGray}{gray}{0.302} 

\makeatletter

\def\@seccntformat#1{%
  \csname the#1\endcsname
  \csname ICAS@secdot@#1\endcsname
  \quad
}
\expandafter\def\csname ICAS@secdot@section\endcsname{.}
\makeatother

\TitlePaper{Coverage-Driven Verification for Safety-by-Design in AI-Based Collision Avoidance Systems
}
\AuthorPaper[1]{Thomas Stefani}
\AuthorPaper[2]{Johann Maximilian Christensen}
\AuthorPaper[2]{Elena Hoemann}
\AuthorPaper[2]{Frank Köster}
\AuthorPaper[2]{Sven Hallerbach}
\affil[1]{%
\textit{Institute for AI Safety and Security}\\
\textit{German Aerospace Center (DLR)}\\
Ulm, Germany%
}
\affil[2]{%
\textit{Institute for AI Safety and Security}\\
\textit{German Aerospace Center (DLR)}\\
Sankt Augustin, Germany%
}

\abstractEnglish{%
Artificial Intelligence (AI) offers significant potential for future aviation systems; however, its integration into safety-critical applications requires compliance with the aviation sector's stringent safety standards.
For AI and Machine Learning (ML)-based systems, the European Union Aviation Safety Agency (EASA) emphasizes the need to demonstrate the representativeness and completeness of the Operational Design Domain (ODD) and the associated data distributions used during development and verification.
Despite this requirement, a structured engineering process for defining target distributions and evaluating representativeness within ODDs remains largely unexplored.
This work presents a method for representativeness assessment of AI/ML constituent ODDs in the context of aviation safety assurance.
Starting from the methodical identification of suitable target distributions, a process flow is proposed that guides developers from ODD definition and parameter distribution modeling to the quantitative assessment and interpretation of coverage results with respect to EASA's learning assurance objectives.
As quantitative measures, the chi-squared goodness-of-fit test is examined and found unsuitable for the large data sets arising in this setting, leading to the adoption of the Kullback--Leibler divergence and Cramér's $V$ for the representativeness assessment.
The method is demonstrated using the example of AI-based airborne collision avoidance, employing experimental data from previous Horizontal Collision Avoidance System (HCAS) and Vertical Collision Avoidance System (VCAS) simulations.
The results illustrate how statistical distribution comparison methods can support the assessment of representativeness for safety-critical AI applications and contribute toward a systematic Safety-by-Design AI engineering process aligned with emerging EASA guidance.
}

\begin{document}

\body  

\section{Introduction}
Current developments in Artificial Intelligence (AI) are transforming industries, promising exponential gains in efficiency, while simultaneously presenting engineers with novel challenges.
Here, aviation is no exception.
Analysts expect annual growth rates of approximately \qty{35}{\percent} for AI-based applications in aviation~\cite{PrecedenceResearch2022}.
This is driving ongoing research into various AI-based safety-critical applications, such as collision avoidance, particularly simplified versions of ACAS~X called HCAS and VCAS~\cite{Julian2019a}.  
The increasing complexity of AI challenges regulators in safety assurance and certification~\cite{Christensen2024a}.
However, the use of any AI-based system in aviation requires that the development process comply with European Union Aviation Safety Agency (EASA) guidelines.
Within the learning assurance framework, EASA emphasizes the need to demonstrate the \enquote{completeness and representativeness of data sets} used for the development and verification of AI/ML constituents~\cite{EUASA2024}.
In particular, the AI/ML constituent Operational Design Domain (ODD) defines the operational conditions and corresponding input space under which the system is intended to operate safely~\cite{EUASA2024}.
Consequently, the verification data used throughout the assurance process must adequately represent the distributions within the defined ODD.
One step in this direction was taken in previous work~\cite{Stefani2024a, Stefani2024b, Christensen2024a}, which focused on deriving scenarios from ODD descriptions using the collision-avoidance use cases HCAS and VCAS.
By integrating HCAS and VCAS into the generic open-source Python library pyCASX~\cite{Christensen2024a}, both vertical and horizontal advisories can be evaluated within the flight simulator FlightGear.
The generated experimental data now provide a foundation for investigating representativeness assessment methods for AI-based collision avoidance systems.
Despite EASA's explicit emphasis on representativeness, it remains unclear how developers should define suitable target distributions for ODD parameters and quantitatively assess whether available verification data adequately represent these distributions.
Existing approaches for coverage measures often rely on geometric coverage metrics or statistical measures for comparing probability distributions.
However, many of these methods suffer from scalability issues in higher-dimensional spaces, or do not directly address the statistical representativeness of the data distribution.
Therefore, this paper presents a method for representativeness assessment of AI/ML constituent ODDs within the context of aviation safety assurance.
A process flow is proposed that guides developers from defining target parameter distributions to evaluating representativeness using statistical distribution comparison methods.
For this purpose, the chi-squared goodness-of-fit test, the Kullback--Leibler (KL) divergence, and Cramér's $V$ are investigated regarding their applicability to ODD representativeness assessment.
Using experimental HCAS and VCAS data, the proposed method demonstrates how representativeness analysis can support a Safety-by-Design AI engineering process while aligning with EASA's emerging AI safety requirements.

The remainder of this paper is structured as follows.
\Cref{sec:soa} reviews the state of the art in ODD coverage assessment for AI-based systems and identifies the gap addressed in this work.
The proposed method is then introduced in \cref{sec:method}, presenting the two-step coverage assessment loop and detailing the representativeness analysis based on statistical distribution comparison.
Building on this, \cref{sec:UseCase} applies the method to the HCAS and VCAS collision avoidance use cases, describing the experimental setup, the definition of target distributions,
and the resulting representativeness assessment.
The findings are discussed in \cref{sec:Disscusion}, with particular focus on the limited suitability of the chi-squared test for large-scale assessment and the complementary roles of the KL divergence and Cramér's $V$.
Finally, \cref{sec:Conclusion} concludes the paper and outlines directions for future work toward a comprehensive coverage-driven verification process.

\section{State of the Art}\label{sec:soa}
The recent progress in AI prompted the EASA to launch an AI Roadmap~\cite{EUASA2020} and to publish a concept paper~\cite{EUASA2024} addressing Level~1 and Level~2 AI applications.
Central to these publications are the notion of the ODD and the learning-assurance framework built on the W-shaped process, which prescribes dedicated steps for the trustworthy development of AI-based systems, including the demonstration of full ODD coverage.
How to translate these objectives into a usable AI engineering process, however, remains only loosely defined~\cite{MLEAPConsortium2024}.
Prior work~\cite{Stefani2024b, Christensen2025} approached this by combining the W-shaped process with the DevOps cycle, yielding a first AI engineering framework for AI certification in aviation, while Werner~\cite{Werner2025} established how to methodically derive the AI/ML constituent Operational Domain (OD) and ODD.
Because the AI/ML constituent ODD defines the input space of the neural network, demonstrating sufficient coverage of this space is essential to the safety argument~\cite{Namiot2024}.
In safety-critical systems, \emph{coverage} has traditionally referred to structural code metrics, but AI-based functions shift the focus toward verifying the ODD itself~\cite{Hirschle2024}.
In line with the EASA objectives, demonstrating ODD coverage rests on two complementary properties: completeness, whether the data span the ODD combinatorially, and representativeness, whether the data follow the distribution it prescribes.
Most existing work addresses completeness.
Weissensteiner~\cite{Weissensteiner2023} defined a high-level ODD coverage process for validating automated driving systems and proposed a sampling method for n-dimensional scenario-parameter distributions, in which an initial k-means clustering is adapted under predefined boundary conditions to require substantially fewer scenarios; the approach strengthens the overall safety argument but does not account for attribute interactions and the resulting high dimensionality.
Using combinatorial testing, Diemert~\cite{Diemert2023} introduced the CACTus framework to systematically model the ODD input space and derive a manageable scenario suite, showing that combinatorial testing together with expert judgment can reduce the parameter space, though without quantifying how representatively that space is populated.
Geometry-based metrics form a further line of work.
Hirschle~\cite{Hirschle2024} examined a convex hull enclosing all samples, which does not credit new samples falling inside the hull and scales poorly in higher dimensions, as well as a per-sample hypersphere metric, which can overestimate coverage and yield values above one depending on the chosen radius.
As a solution, Hirschle proposed two new metrics, of which a kernel-density-estimation metric showed promising scalability and the ability to localize low-coverage regions.
Common to all these approaches is that they quantify how much of the input space is covered, rather than whether the data within it follow the distribution prescribed by the ODD.
The latter facet is addressed most directly by the joint EASA and Collins Aerospace ForMuLA report~\cite{ForMuLA}, which demonstrated a representativeness assessment of an ML constituent ODD within a learning-assurance toolchain.
For a remaining-useful-life prediction use case, the durations of individual flight regimes were compared against the Gaussian target distributions prescribed in the ODD using goodness-of-fit tests, with the chi-squared test as the primary example alongside the Kolmogorov–Smirnov, Lilliefors, and Anderson–Darling tests.
There, representativeness was reduced to a binary pass/fail verdict at a fixed significance level, and the authors themselves reported that data may be rejected even when they visually match the target, that raising the significance level alters the verdict but demands further justification, and that small or empty expected bin counts together with a limited dataset size reduce the reliability of the test.
While this confirms the feasibility of statistical representativeness assessment in an EASA context, it was demonstrated on a single low-dimensional feature and relied on a single significance-test outcome, leaving open how representativeness can be quantified robustly across a high-dimensional ODD.

Despite these advances in quantifying ODD sub-spaces, the literature still lacks a comprehensive coverage process that satisfies EASA's coupled requirements for data representativeness and completeness, both of which are essential for the formal certification of AI-based systems.

\section{Method for ODD Coverage Assessment}\label{sec:method}
The assurance of safety-critical AI-based systems requires methods that enable a systematic and trustworthy verification of their behavior.
In aviation, where AI/ML applications must comply with stringent safety requirements, regulatory authorities have introduced guidance to support the certification and assurance of such systems.
In particular, the EASA defines several anticipated means of compliance (MOC) addressing the coverage of the ODD, including requirements for demonstrating completeness and representativeness of the data and parameter space~\cite{EUASA2024}.
However, despite these objectives and their corresponding MOCs being formulated at a high level, the practical demonstration of sufficient ODD coverage remains largely unresolved.
Existing guidance does not specify how completeness and representativeness should be quantified, nor which metrics and processes are adequate for demonstrating coverage in high-dimensional ODDs.
In other works~\cite{Stefani2026}, a method for demonstrating completeness is proposed.
Still, there is currently a lack of a generic and scalable method that accounts for the representativeness of the data and supports the verification of AI/ML constituents while aligning with EASA's Safety-by-Design and learning assurance principles.

\subsection{Method Overview}\label{sec:overview}
\begin{figure}[htbp] 
    \centering
    \includegraphics[width=\linewidth]{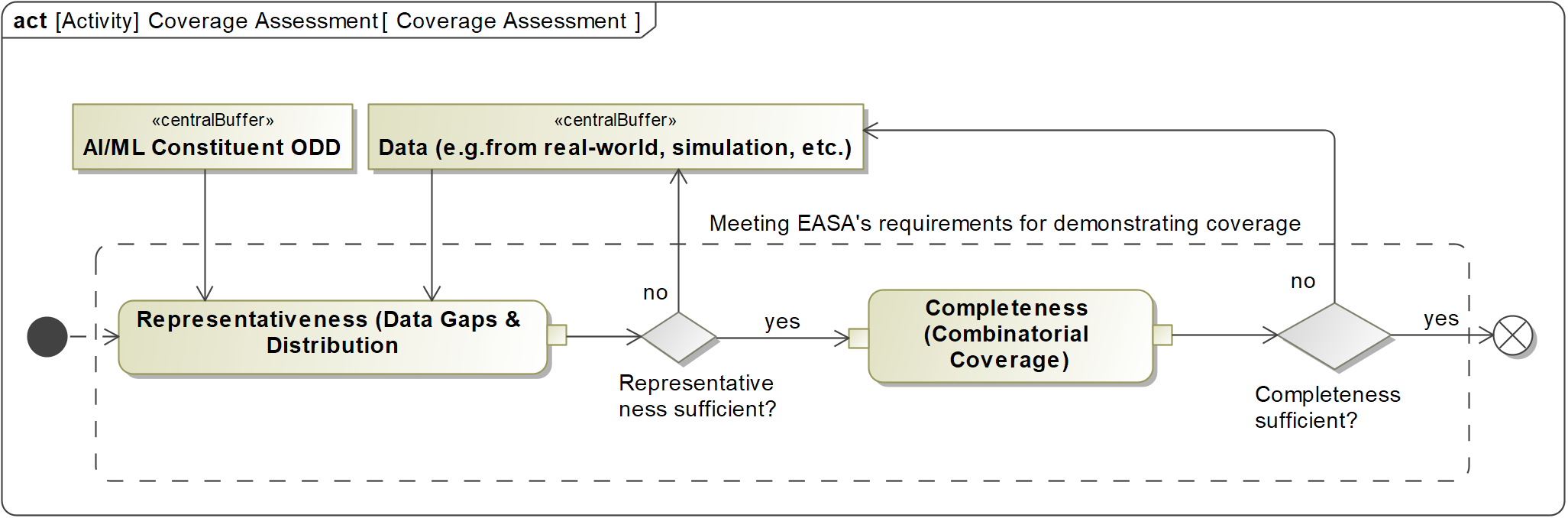}
    \caption{Activity Diagram for the Coverage Assessment loop aligning with EASA's MOC DM-08 through a two-stage process consisting of demonstrating data representativeness and completeness.}
    \label{fig:MBSE_1} 
\end{figure}

The method proposed in this work operationalizes EASA's coverage assessment objectives through a structured two-step process, as illustrated in \cref{fig:MBSE_1}.
The method takes two inputs: the formally defined AI/ML Constituent ODD and a dataset, which in this case are simulated near mid-air collision scenarios.
The use case is introduced in more detail in \cref{sec:UseCase}.
Together, these inputs drive an iterative assessment loop that evaluates whether the dataset sufficiently covers the operational parameter space.
The assessment proceeds sequentially.
First, a representativeness analysis examines whether the dataset adequately reflects the distribution and density of the ODD parameter space, identifying gaps and underrepresented regions.
If representativeness is found to be insufficient, the loop feeds back, requiring dataset augmentation before proceeding further.
Only once representativeness is established does the process advance to the second step: a completeness analysis, which verifies that all relevant combinations of ODD parameters are covered to the required degree~\cite{Stefani2026}.
If both criteria are satisfied, coverage is deemed sufficient, and the assessment concludes.
If completeness is found lacking despite adequate representativeness, the loop again iterates, targeting the specific parameter combinations that remain uncovered.
This sequential structure follows a logical dependency: parameter distribution must be evaluated before parameter combinations.
Because addressing any gaps found in the underlying distribution requires adding new data, doing so directly alters the dataset's combinatorial mix, meaning any completeness check performed beforehand would be invalidated.

\subsection{Data Representativeness}\label{sec:representativeness}
%
%
Demonstrating data representativeness is a central aspect across several of EASA's
anticipated objectives~\cite{EUASA2024}.
At the data-management requirements stage, Objective~DA-04 requires the applicant to capture the data quality requirements for all training, validation, and test data, explicitly including the completeness and representativeness of the data sets and the traceability of their origin.
The dedicated representativeness means of compliance, Anticipated MOC~DM-07-2, characterizes representativeness as the property that the data are sampled according to the distribution prescribed for the input space, and anticipates statistical goodness-of-fit methods for assessing operating parameters~\cite{EUASA2024}.
Finally, the data-verification step of Objective~DM-08 confirms that the test data set covers the whole ODD with the necessary level of completeness and representativeness. The method proposed here operationalizes this chain for the representativeness property: starting from the requirements captured under DA-04, it provides the quantitative assessment anticipated by DM-07-2 and produces the evidence consolidated at the DM-08 verification step~\cite{EUASA2024}.
This work addresses the distributional aspect of DM-07-2 at the level of the individual ODD parameters: for each parameter, the observed data are compared against the univariate distribution prescribed for it, so that representativeness is established parameter by parameter.
This parameter-level representativeness is one of the two coupled properties on which EASA's coverage demonstration rests; its counterpart, the pairwise combinatorial coverage of the ODD parameters anticipated by Objective LM-16, is the object of the complementary completeness pillar and is treated separately~\cite{Stefani2026}.
Two aspects therefore lie outside the scope of the present contribution: the sampling independence between data sets (Anticipated MOC DM-07-5), which is assumed rather than assessed, and any characterization of the joint ODD density beyond these two coupled properties, which is not treated in this work.
Consequently, the process flow for demonstrating representativeness is defined as follows.
The AI/ML constituent ODD and the generated simulation data are assumed as given inputs.
Based on the ODD definition, the first step consists of identifying the target distribution for each relevant ODD parameter.
Once the target distributions are defined, a representativeness assessment is performed to quantify how well the available dataset represents the expected operational parameter space.
To this end, different metrics and divergences are investigated, including the KL divergence, the chi-squared test, and Cramér's V.
The objective of the representativeness step is therefore to identify underrepresented regions and distribution mismatches within the ODD before proceeding to the completeness assessment.
Subsequent coverage analyses yield meaningful assurance regarding AI/ML constituent behavior only if the dataset adequately represents the intended ODD.

\subsubsection{Target Distribution}\label{sec:target_dist}
\begin{figure}[htbp]
\centering
\begin{tikzpicture}[node distance=1.8cm]


\node (start) [decision] {Real-world data\\available?};

\node (odd) [data, left=2cm of start] {ODD parameter};

\node (out1) [block, right=2.5cm of start] {Use real\\data};

\node (dec2) [decision, below of=start, yshift = -1cm] {Simulation data\\available?};

\node (out2) [block, right=2.5cm of dec2] {Use simulation\\data};

\node (dec3) [decision, below of=dec2, yshift=-2cm] {Physically motivated assumption\\possible (e.g., uniform, normal)?};

\node (out3) [block, right=1cm of dec3] {Use statistical\\assumption};

\node (dec4) [decision, below of=dec3, yshift=-2.5cm] {Expert knowledge or domain\\standards available?};

\node (out4) [block, right=1.4cm of dec4] {Use expert\\knowledge};

\node (flag) [block, below of=dec4, yshift=-1.2cm] {Flag parameter: distribution\\undefined---next steps tbd.};

\node (merge) [coordinate, below of=flag, yshift=0.7cm] {};

\node (doc) [block, below of=merge, yshift=1.1cm] {Document source and rationale\\for traceability (EASA Objective DA-04)};

\node (cov) [block, below of=doc, yshift=0.2cm] {Apply coverage assessment};

\node (data) [data, left=2.5cm of cov] {Data};

\node (ver) [block, below of=cov, yshift=0.2cm] {Representativeness verified\\or gap flagged for mitigation};

\path [line] (odd) -- (start);
\path [line] (start) -- node[left, yshift=-0.2cm]{No} (dec2);
\path [line] (dec2) -- node[left, yshift=-0.2cm]{No} (dec3);
\path [line] (dec3) -- node[left, yshift=-0.2cm]{No} (dec4);
\path [line] (dec4) -- node[left, yshift=-0.2cm]{No} (flag);
\path [line] (flag) -- (doc);
\path [line] (doc) -- (cov);
\path [line] (data) -- (cov);
\path [line] (cov) -- (ver);

\path [line] (start) -- node[above]{Yes} (out1);
\path [line] (dec2) -- node[above]{Yes} (out2);
\path [line] (dec3) -- node[above]{Yes} (out3);
\path [line] (dec4) -- node[above]{Yes} (out4);

\path [line] (out1) -- ++(2.0,0) |- (doc);
\path [line] (out2) -- ++(2.0,0) |- (doc);
\path [line] (out3) -- ++(2.0,0) |- (doc);
\path [line] (out4) -- ++(2.0,0) |- (doc);

\end{tikzpicture}
\caption{
Generic process flow for identifying the target distributions for the ODD parameters that are subsequently used for the coverage assessment. Starting from real-world data, over simulation data, physically motivated assumptions, to expert knowledge. }
\label{fig:target}
\end{figure}
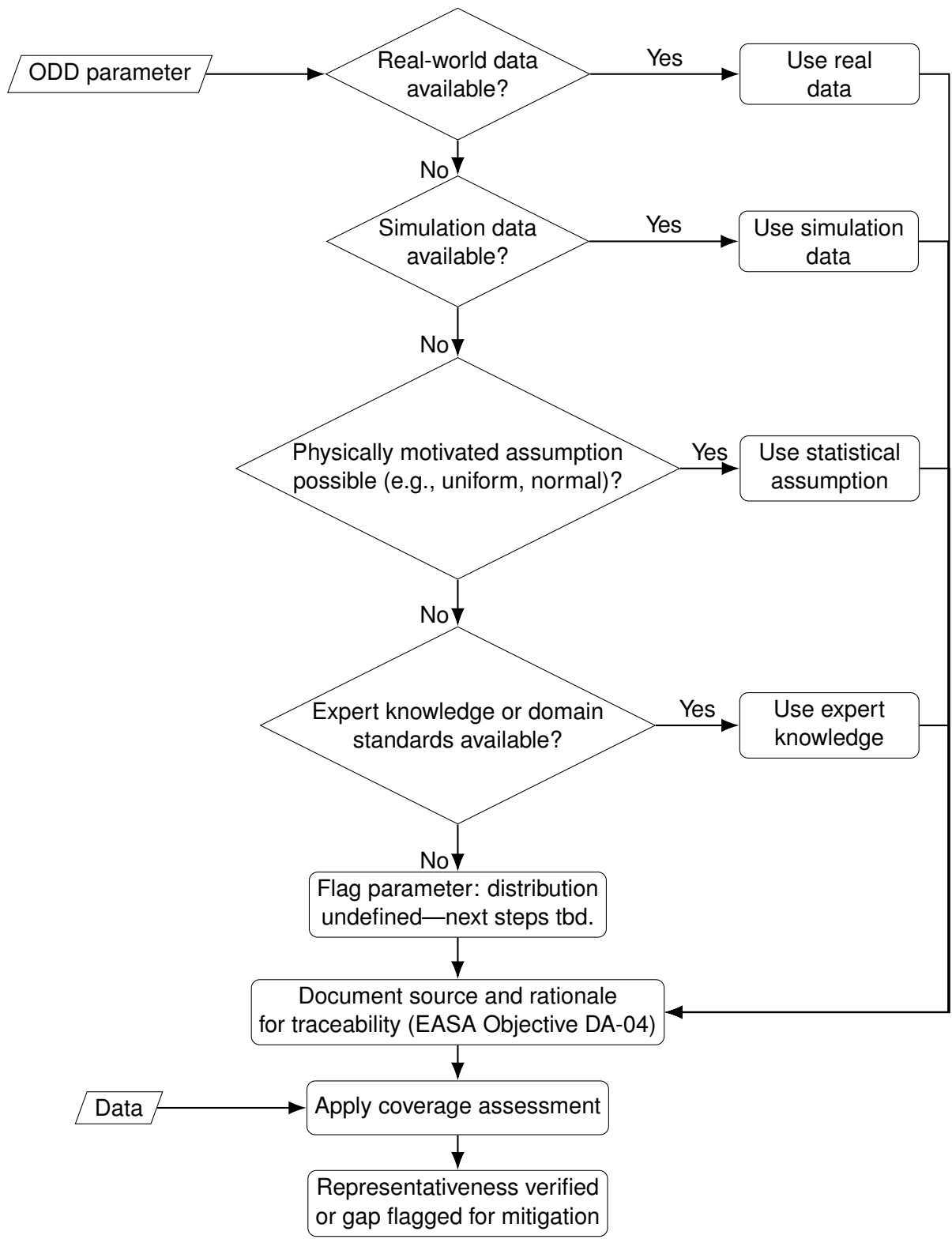

To assess representativeness, a target distribution must first be defined for each parameter within the AI/ML constituent ODD.
\cref{fig:target} illustrates a structured process flow to guide engineers in systematically deriving suitable target distributions from available data sources, statistical assumptions, or expert knowledge.
Starting from a formally defined ODD parameter, the method first evaluates whether representative real-world data are available.
If such data exist, they are directly used to derive the target distribution.
Otherwise, the process investigates whether simulation data are available from representative reference simulations.
If neither real-world nor simulation data are available, the method evaluates whether a physically motivated statistical assumption can reasonably describe the parameter distribution, for example, using uniform or normal distributions.
If this is not feasible, expert knowledge  or existing domain standards are considered as alternative sources for defining the target distribution.
In cases where none of these information sources are available, the parameter is flagged as undefined and requires further investigation.
Regardless of the selected source, the rationale and origin of the chosen target distributions are documented to ensure traceability in accordance with EASA MOC~DM-07-2.
The resulting target distributions are subsequently used within the coverage assessment process to evaluate representativeness and identify potential coverage gaps requiring mitigation.

\subsubsection{Representativeness via Kullback--Leibler Divergence}\label{sec:cov_metrics1}

The KL divergence, or relative entropy, is an asymmetric statistical measure used to quantify the difference between two probability distributions.
In the context of data representativeness, it can be used to evaluate how well the observed test data distribution, typically defined as $Q(x)$, matches a target or expected distribution $P(x)$.
For a continuous parameter space, the divergence $D_\mathrm{KL}$ is defined via integrating over the parameterspace~\cite{Kullback1951}:
\begin{equation}
    D_\mathrm{KL}(P \parallel Q) = \int_{-\infty}^{\infty} p(x) \log \left( \frac{p(x)}{q(x)} \right) \mathrm{d}x\text{.}
\end{equation}
To compute a continuous KL divergence, the exact probability density functions (PDFs) must be known.
However, since data or simulation outputs are collections of discrete samples, it is necessary to estimate these continuous curves using techniques such as kernel density estimation (KDE).
Further, if multiple parameters are involved, calculating this integral numerically becomes computationally expensive.
Therefore, we focus explicitly on the discrete formulation:
\begin{equation}
D_\mathrm{KL}(P \parallel Q) = \sum_{x \in \mathcal{X}} P(x) \log \left( \frac{P(x)}{Q(x)} \right)\text{.}
\end{equation}
Because the discrete divergence is unbounded whenever a bin carries target mass but contains no observed samples ($P(x) > 0$, $Q(x) = 0$), the observed distribution is regularized prior to evaluation: each empty bin is assigned a floor probability $\varepsilon = 10^{-6}$, and $Q$ is subsequently renormalized.
This keeps $D_\mathrm{KL}$ finite while confining the divergence of unpopulated bins to a defined, bounded contribution.
As the magnitude of these empty-bin contributions depends on the choice of $\varepsilon$, a single value is fixed across all parameters and documented here to ensure reproducibility in accordance with EASA MOC~DM-07-2.
A lower divergence indicates that the generated or collected data is more representative of the intended operational environment.
The KL divergence is governed by three key properties, namely its non-negativity ($D_\mathrm{KL}(P \parallel Q) \ge 0$), its identity ($D_\mathrm{KL}(P \parallel Q) = 0 \iff P = Q$), and its asymmetry ($D_\mathrm{KL}(P \parallel Q) \neq D_\mathrm{KL}(Q \parallel P)$), making it an adequate choice to assess the representativeness because it quantifies information loss when approximating a true distribution while prioritizing a specific direction of error.
Since EASA's anticipated means of compliance, such as MOC~DM-07-2, require evidence for the \emph{representativeness of data sets} without prescribing concrete quantitative thresholds, heuristic interpretation ranges are proposed in \cref{tab:odd_thresholds}.
These ranges are intended to support the coverage argumentation process and provide a practical guideline for evaluating whether the sampled data adequately represent the AI/ML constituent ODD.
The proposed thresholds are not regulatory limits, but engineering-oriented guidance values derived from common interpretations of statistical divergence measures in distribution comparison and dataset shift analysis~\cite{Belov2011, Bonnici2024, Kvalseth2017}.
\begin{table}[htb]
\centering
\caption{Interpretation of KL divergence values demonstrating parameter representativeness (EASA MOC~DM-07-2)~\cite{Belov2011, Bonnici2024, Kvalseth2017}. Lower $D_\mathrm{KL}$ indicates a better fit between the two distributions.}
\label{tab:odd_thresholds}
\begin{tabular}{rll}
\toprule
$D_\mathrm{KL}$ Ranges & Interpretation & Coverage Assessment Implication \\
\midrule
$0 \le D_\mathrm{KL} < 0.05$ & Excellent & Strong evidence of representativeness \\
$0.05 \le D_\mathrm{KL} < 0.15$ & Good & Generally acceptable with rationale \\
$0.15 \le D_\mathrm{KL} < 0.30$ & Moderate & Mitigation or targeted sampling recommended \\
$0.30 \le D_\mathrm{KL} < 0.50$ & Poor & Coverage argument weakened\\
$D_\mathrm{KL} \ge 0.50$ & Fail & Representativeness claim at risk \\
\bottomrule
\end{tabular}
\end{table}

\subsubsection{Representativeness via Chi-Squared}\label{sec:cov_metrics2}
Anticipated MOC~DM-07-2 names the chi-squared test, alongside the Z- and Kolmogorov--Smirnov tests, among the goodness-of-fit methods anticipated for assessing the representativeness of operating parameters~\cite{EUASA2024}.
The chi-squared goodness-of-fit test is a non-parametric hypothesis test used to assess whether an observed sample distribution is consistent with a fitted target distribution~\cite{Pearson1900}.
In the context of ODD coverage verification, it operationalizes the representativeness requirement: given a target distribution specified for an ODD parameter, the test quantifies whether the collected scenario data reflect that distribution sufficiently well.
The test statistic partitions the parameter range into $k$ non-overlapping bins and compares the observed counts against the expected counts under the target distribution:
\begin{equation}
\chi^2 = \sum_{i=1}^{k} \frac{(Q_i - n\cdot p_i)^2}{n\cdot p_i}\text{,}\label{formula:1}
\end{equation}
where $Q_i$ is the observed count in bin $i$ and $n\cdot p_i$ the corresponding expected count, with $n$ the total number of samples and $p_i$ the probability mass assigned to bin $i$ by the target distribution.
Under the null hypothesis $H_0$ that the observed data are drawn from the target distribution, the statistic follows a chi-squared distribution with $\nu = k - 1$ degrees of freedom.
$H_0$ is rejected at significance level $\alpha$ if $\chi^2 > \chi^2_{\alpha, \nu}$; a failure to reject, that is, a $p$-value exceeding $\alpha$, is taken as evidence that the data are consistent with the target and thereby supports the representativeness claim.
Following common practice, a conventional fixed level of $\alpha = 0.05$ is used as the baseline criterion~\cite{KennedyShaffer2019}.
A practical prerequisite of the test is that the expected counts be sufficiently large, commonly $n\cdot p_i \geq 5$ for each bin, to ensure the validity of the chi-squared approximation~\cite{Fisher1992}.
This couples the choice of bin granularity to the available dataset size: finer bins increase the sensitivity to local distributional deviations but require proportionally more data.
Its applicability to any target distribution---uniform, normal, or empirically defined---further makes the test well suited to the heterogeneous nature of ODD parameters.
A decisive limitation, however, is that the $\chi^2$ statistic grows with the sample size $n$, so that for large data sets even negligible deviations are flagged as significant.
This motivates the sample-size-normalized effect-size formulation introduced in the following subsection.

\subsubsection{Representativeness via Cramér's V}\label{sec:cov_metrics3}
The chi-squared goodness-of-fit test of \cref{sec:cov_metrics2} yields a binary verdict on the \emph{existence} of a distributional difference, but its statistic scales with the sample size $n$: as $n$ grows, even practically negligible deviations become statistically significant, so the test tends to reject $H_0$ for any sufficiently large data set~\cite{KennedyShaffer2019}.
To assess representativeness in terms of the \emph{magnitude} of the deviation, the $\chi^2$ statistic from \cref{formula:1} is converted into an effect-size measure.
Cramér's $V$~\cite{Cramer2016} normalizes a $\chi^2$ statistic by the sample size and the table dimensionality,
\begin{equation}
V = \sqrt{\frac{\chi^2}{n\,(q-1)}}, \qquad q = \min(r, c).
\end{equation}
where $r$ and $c$ denote the number of rows and columns of the underlying contingency table.
For the goodness-of-fit comparison, the smaller dimension is $q = 2$ (the observed against the target distribution over $k$ bins), so that the measure applied to the statistic of \cref{formula:1} is
\begin{equation}
V = \sqrt{\frac{\chi^2}{n}}.
\end{equation}
This quantity is equivalent to Cohen's effect-size measure $w$ for a chi-squared goodness-of-fit test~\cite{Cohen1988}.
By normalizing $\chi^2$ by the sample size, $V$ removes the large-sample inflation that drives the bare $\chi^2$ test toward rejection, while reusing the identical binning and target-distribution inputs already required for the $\chi^2$ computation.
It satisfies $V \ge 0$, with $V = 0$ if and only if the observed and target distribution coincide, and larger values indicating progressively stronger deviation. Unlike the association measure of a full contingency table, the goodness-of-fit effect size is not bounded above by one: for equiprobable binning, its maximum is $\sqrt{k-1}$, so $V$ must be interpreted relative to a fixed bin resolution.
Being constructed from a different principle than the information-theoretic KL divergence, $V$ provides a construction-independent second indicator of representativeness, allowing the two measures to be used as a mutual cross-check.
As with the KL divergence, EASA's anticipated means of compliance do not prescribe quantitative thresholds for $V$.
The interpretation ranges in \cref{tab:cramersv}; therefore, follow Cohen's conventional effect-size magnitudes for $\mathrm{df}=1$~\cite{Cohen1988}, adapted as engineering-oriented guidance rather than regulatory limits.
A stricter pass threshold may furthermore be assigned to parameters of higher safety relevance~\cite{Gibbs2002}.
\begin{table}[h]
\centering
\caption{Interpretation of Cramér's $V$ values for the $2 \times k$ goodness-of-fit framing, demonstrating parameter representativeness (EASA MOC~DM-07-2).
The thresholds are adopted from Cohen~\cite{Cohen1988} and translated into finer granularity.}
\label{tab:cramersv}
\begin{tabular}{@{}rll@{}}
\toprule
$V$ Ranges & Interpretation & Coverage Assessment Implication \\
\midrule
$0 \le V < 0.10$        & Excellent & Strong evidence of representativeness \\
$0.10 \le V < 0.20$ & Good       & Generally acceptable with rationale \\
$0.20 \le V < 0.30$ & Moderate    & Mitigation or targeted sampling recommended \\
$0.30 \le V < 0.40$ & Poor    & Coverage argument weakened \\
$V \geq 0.40$     & Fail     & Representativeness claim at risk \\
\bottomrule
\end{tabular}
\end{table}

\section{Case Study: Horizontal and Vertical AI-Based Collision Avoidance}\label{sec:UseCase}
The following case study applies the representativeness method from \cref{sec:method} to a concrete aviation AI application: an AI-based collision avoidance system comprising a HCAS and a VCAS~\cite{Julian2019a}.
This use case was selected for two reasons.
First, it represents a well-defined, safety-critical AI/ML constituent with a clearly bounded ODD, making it well-suited for demonstrating the proposed coverage assessment.
Second, both systems have been tested in prior work~\cite{Stefani2024a, Stefani2024b, Christensen2024a}, providing the simulation infrastructure and scenario data necessary to populate the ODD.
VCAS and HCAS are neural-network-based proof-of-concept implementations of the ACAS X family of collision avoidance standards, addressing the vertical and horizontal separation dimensions, respectively~\cite{Julian2019a}.

\subsection{ODD Definition}\label{sec:ODDdef}

HCAS issues horizontal advisories based on seven state variables: the range to intruder $\rho$, the bearing angle $\theta$, the relative heading $\psi$, the ownship speed $v_\mathrm{own}$, the intruder speed $v_\mathrm{int}$, the time to closest point of approach (CPA) $\tau$, and the previous advisory $s_\mathrm{adv}$.
VCAS extends this logic to the vertical plane, issuing vertical advisories---such as climb or descend commands---based on five state variables which are listed in~\cite{Julian2019a}: the relative altitude between the ownship and the intruder $h$, the own climb/descent rate $\dot{h}_\mathrm{own}$, the intruder's climb/descent rate $\dot{h}_\mathrm{int}$, the time to CPA $\tau$, and the previously issued advisory $s_\mathrm{adv}$.
While a traditional ODD typically contains a broader set of environmental attributes, such as scenery and environmental conditions, the focus here is on the immediate input space of the neural networks.
Consequently, for the purpose of this analysis, these state variables define the AI/ML constituent ODD for each system, as summarized in \cref{tab:HCAS_ODD} and \cref{tab:VCAS_ODD}.
While our parameter values are adapted from Kochenderfer et al.~\cite{Julian2019}, they vary slightly from those used in other prior studies~\cite{Christensen2024a, Christensen2024, Stefani2024a, Stefani2024b, Stefani2026}.
This adjustment is due to differences in scenario settings, as the primary objective of this work is to demonstrate the underlying process flow rather than to focus strictly on numerical precision.
For reasons of simplification, for both HCAS and VCAS, the previous advisory is not considered in the ODD.
\begin{table}[htb]
    \caption{Derived AI/ML constituent $ODD_{HCAS}$ from the HCAS State Variables~\cite{Julian2019}. The values are oriented on the GitHub release at \url{https://github.com/sisl/HorizontalCAS} but differ based on previous work~\cite{Christensen2024a, Christensen2024, Stefani2024a, Stefani2024b} where \(v_\mathrm{own}\) was kept the same throughout the executed simulations and therefore remained at a constant value.}
    \label{tab:HCAS_ODD}
    \centering
    \begin{tabular}{lll}
        \toprule
        Variable           & Description       & Range                                 \\
        \midrule
        \(\rho\)           & Range to intruder & [\qty{0}{\foot}, \qty{56000}{\foot}]  \\
        \(\theta\)         & Bearing angle     & [\ang{-180}, \ang{180}]               \\
        \(\psi\)           & Relative heading  & [\ang{-180}, \ang{180}]               \\
        \(v_\mathrm{own}\) & Ownship speed     & \qty{865}{\foot\per\second}           \\
        \(v_\mathrm{int}\) & Intruder speed    & [\qty{845}{\foot\per\second}, \qty{868}{\foot\per\second}]         \\
        \(\tau\)           & Time to CPA       & [\qty{0}{\second}, \qty{55}{\second}] \\
        \bottomrule
    \end{tabular}
\end{table}
\begin{table}[htb]
    \caption{Derived AI/ML constituent $ODD_{VCAS}$ from the VCAS State Variables~\cite{Julian2019}. The values are oriented on the GitHub release at \url{https://github.com/sisl/VerticalCAS} but differ based on previous work~\cite{Christensen2024a, Christensen2024, Stefani2024a, Stefani2024b} where only the ownship avoided the collision, therefore \(\dot{h}_\mathrm{int}\) is set to $0$. }
    \label{tab:VCAS_ODD}
    \centering
    \begin{tabular}{lll}
        \toprule
        Variable                 & Description          & Range                                                       \\
        \midrule
        \(h\)                    & Relative altitude    & [\qty{-2800}{\foot}, \qty{2800}{\foot}]                     \\
        \(\dot{h}_\mathrm{own}\) & Ownship vert.\ rate  & [\qty{-3200}{\foot\per\minute}, \qty{3200}{\foot\per\minute}] \\
        \(\dot{h}_\mathrm{int}\) & Intruder vert.\ rate & [\qty{0}{\foot\per\minute}] \\
        \(\tau{}\)               & Time to CPA          & [\qty{0}{\second}, \qty{40}{\second}]                       \\
        \bottomrule
    \end{tabular}
\end{table}
%

\subsection{Simulation Setup and Data Collection}\label{sec:setup}
\begin{figure}[htb]
\centering
\includegraphics[width=0.8\textwidth]{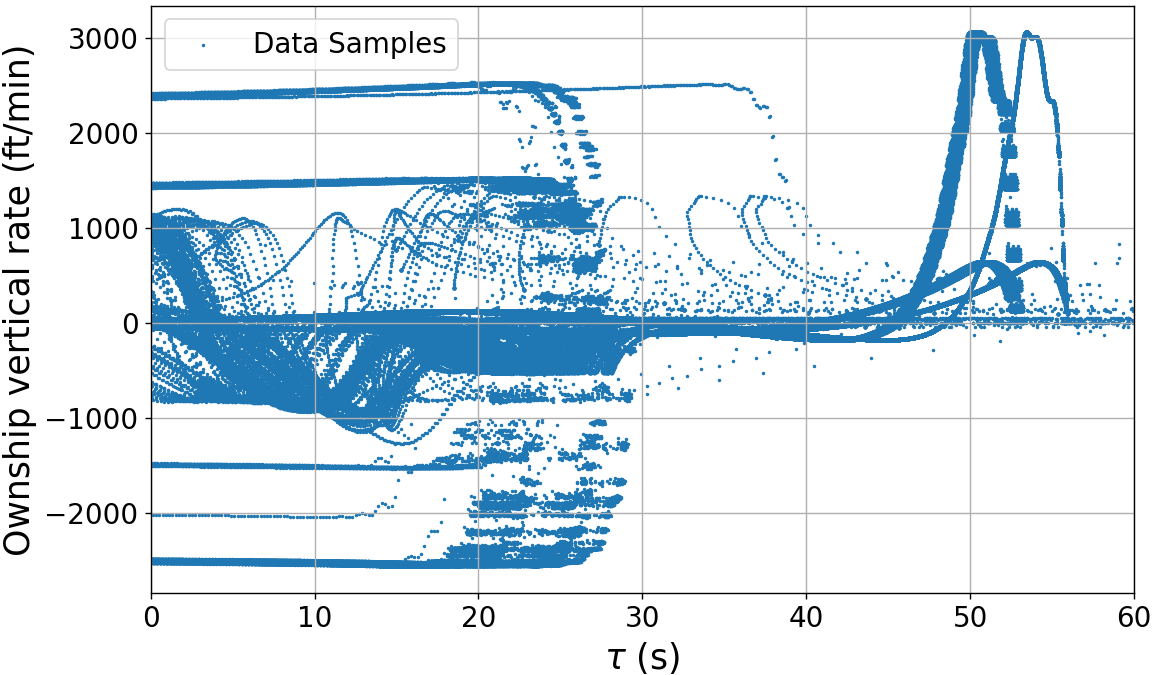}
\caption{Excerpt from the executed collision avoidance scenarios. The blue dots indicate the available data for the ownship's vertical rate over the time to CPA $\tau$. At approximately \qty{30}{\second} to CPA the ownship performs an avoidance maneuver through either climbing or descending.}
\label{fig:h_dot_down_plot}
\end{figure}

The data used for the coverage assessment was obtained from previously executed HCAS and VCAS simulation experiments presented in earlier work~\cite{Stefani2024a, Stefani2024b, Christensen2024a}.
The scenarios were executed in a simulation environment using a highly automated framework to generate randomized trajectories beginning approximately \qty{55}{\second} before CPA.
The resolution advisories provided by HCAS and VCAS  were automatically translated into ownship avoidance maneuvers via an auto-avoid function, at approximately \qty{30}{\second} to CPA.
The simulation outputs, containing the corresponding state variables for either VCAS or HCAS, were collected for future analysis.
\cref{fig:h_dot_down_plot} shows an example excerpt from the VCAS dataset, illustrating the ownship vertical rate $\dot{h}_\mathrm{own}$ plotted over time to CPA ($\tau$).
The blue dots indicate the ownship's climb and descent maneuver issued by VCAS to avoid a potential near mid-air collision between an intruder aircraft and the ownship.
The different reached rates reflect either a strong climb, a climb, a clear of conflict, a descent, or a strong descent.
The original experiments were not conducted with the objective of demonstrating representativeness or ODD coverage.
Instead, existing data are used in this work as an exemplary basis to illustrate how a representativeness assessment process for AI/ML constituent ODD coverage can be performed.
For both HCAS and VCAS, a CSV file was used to store 1.97 million rows of state variables extracted from simulations.

\subsection{Coverage Results for Representativeness}\label{sec:results}
\begin{figure}[htb]
     \centering
     \begin{subfigure}[b]{0.49\textwidth}
         \centering
         \includegraphics[width=\textwidth]{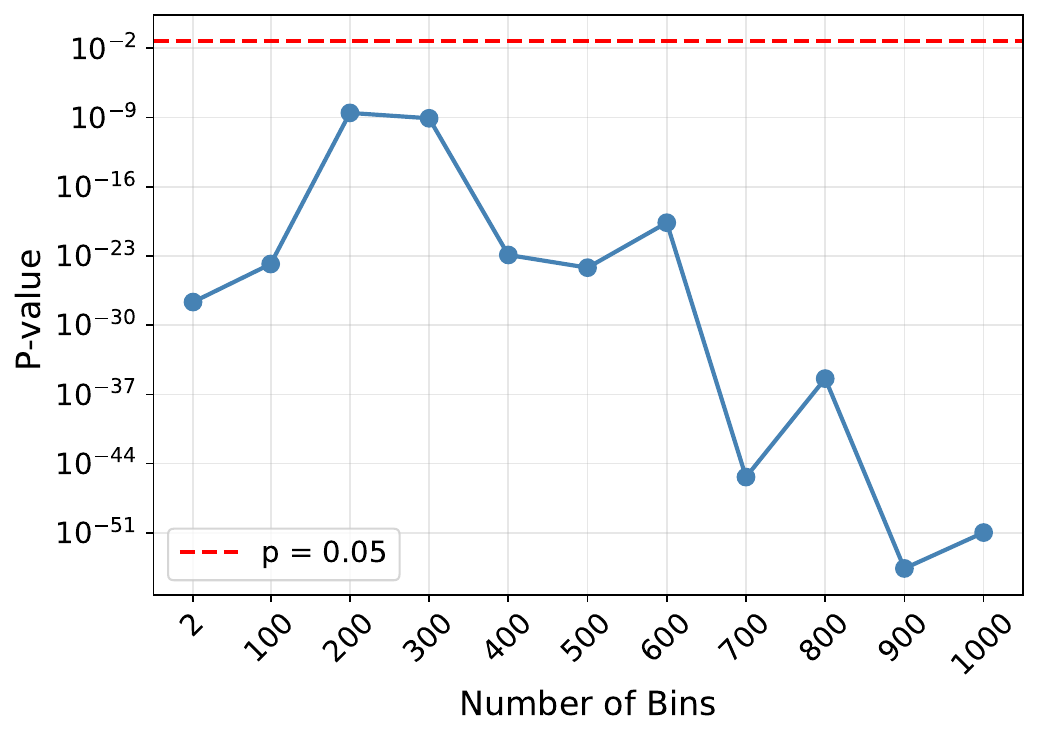}
         \caption{$\chi^2$ sensitivity analysis for the $ODD_{VCAS}$ parameter $\tau$ regarding number of bins. The red dashed line highlights the minimum $p$-value to be reached.  Lower $p$-values indicate stronger deviation.}
         \label{fig:chi_2_sensitivity}
     \end{subfigure}
     \hfill 
     \begin{subfigure}[b]{0.49\textwidth}
         \centering
         \includegraphics[width=\textwidth]{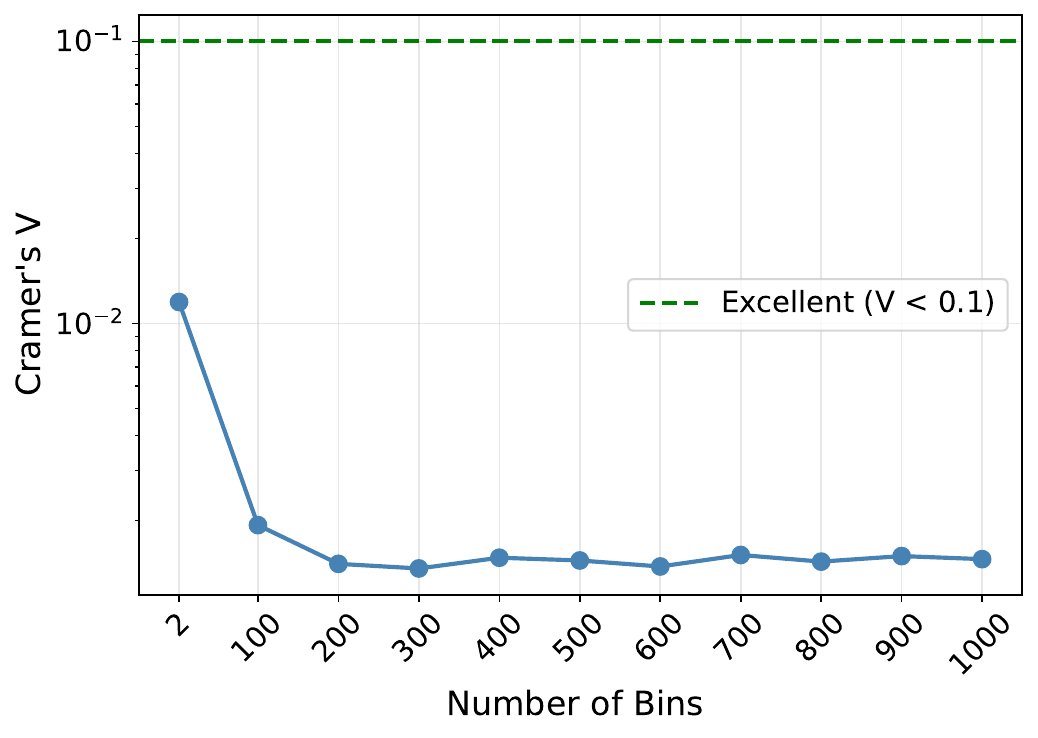}
         \caption{Cramér's $V$ sensitivity analysis for the $ODD_{VCAS}$ parameter $\tau$ regarding bin count. The green line shows the maximum value for \emph{excellent} fit. Consequently, lower $V$-values indicate better alignment.}
         \label{fig:V_sensitivity}
     \end{subfigure}
     \caption{Sensitivity of the $\chi^2$ test and Cramér's~$V$ to the number of bins for the $ODD_{VCAS}$ parameter~$\tau$ which has an approximately uniform distribution.}
     \label{fig:sensitivity_analysis}
\end{figure}

As described in \cref{fig:target}, the proposed process flow for identifying the target distribution was followed.
However, obtaining representative real-world operational data or generating dedicated simulation data is often time-intensive and may not be feasible during early methodological investigations.
Since the focus of this paper is the exemplary demonstration of the proposed representativeness assessment method, the target distributions for the selected ODD parameters were defined based on expert knowledge and engineering assumptions.
To ensure traceability in accordance with EASA MOC~DM-07-2 objectives, the rationale and source for each assumption are documented.
An excerpt for two parameters is given in \cref{tab:target_distribution_rationale}.
\begin{table}[htb]
\centering
\caption{Exemplary documentation of target distribution assumptions for ODD parameters in accordance with EASA MOC~DM-07-2 traceability objectives.}
\label{tab:target_distribution_rationale}
\begin{tabular}{p{2.5cm} p{2.0cm} p{9.8cm}}
\hline
{Parameter} & {Target } & {Rationale / Source} \\
{} & { Distribution} &  \\
\hline
$ODD_{VCAS}$: $\dot{h}_{own}$ 
& Normal distribution 
& The ownship vertical rate is assumed to follow a normal distribution based on expert knowledge of typical aircraft flight behavior. During nominal operations, aircraft predominantly operate around stable climb or descent rates, while extreme vertical rates occur less frequently. The normal distribution, therefore, reflects the expected concentration around standard operational conditions while still accounting for rare deviations. \\
$ODD_{HCAS}$: $\tau$ 
& Uniform distribution 
& In approximately unaccelerated encounters, time to CPA ($\tau$) decays at a constant ratio with elapsed time. Recording potential encounters at a fixed sampling rate over a set window (e.g., from 55 to 0 seconds) therefore yields an approximately uniform distribution.\\
\hline
\end{tabular}
\end{table}
Before presenting the coverage results, we justify why only the KL divergence and Cramér's~$V$ are considered in the coverage report.
The Pearson $\chi^2$ test, although one of the candidate metrics for demonstrating representativeness introduced in \cref{sec:method}, did not yield interpretable results.
To isolate the cause, we conducted a bin-count sensitivity analysis on the VCAS parameter $\tau$.
This parameter was chosen deliberately since the distribution was approximately uniform based on the scenario setting.
Therefore, a representative sample should \emph{not} be rejected ($p>0.05$) and should exhibit negligible association.
\cref{fig:sensitivity_analysis} shows the outcome of the sensitivity analysis.
As shown in \cref{fig:chi_2_sensitivity}, the $\chi^2$ $p$-value remains many orders of magnitude below the \num{0.05} significance level for every bin count and deteriorates further as the number of bins increases.
This is the expected consequence of the large-sample inflation of the $\chi^2$ statistic: at the encounter-set sample sizes used here, statistical significance is driven by sample size and granularity, which renders the $p$-value unusable as a representativeness criterion.
Cramér 's~$ V$, which normalizes the same $\chi^2$ statistic into a bounded effect size, behaves as required (see \cref{fig:V_sensitivity}).
It remains below the negligible-association threshold ($V<0.1$) across the entire sweep and stabilizes with increasing bin counts.
The magnitude of the distributional difference is thus recovered robustly and independently of the binning choice.
We therefore retain $\chi^2$ solely as the documented basis for this decision and report its normalized effect size, Cramér's~$V$, alongside the KL divergence in \cref{tab:vcas_coverage} and \cref{tab:hcas_coverage}, providing the traceable rationale required by MOC~DM-07-2.
\cref{tab:vcas_coverage} and \cref{tab:hcas_coverage} report the ODD coverage results for the VCAS and HCAS input parameters, evaluated with the KL divergence $D_\mathrm{KL}$ and Cramér's $V$, the two measures retained following the sensitivity analysis in \cref{fig:sensitivity_analysis}.
Range utilization denotes the ratio between the interval spanned by the collected data and the defined ODD range of the respective parameter, while bin coverage quantifies how completely the bins within that range are populated; a bin coverage of \qty{100}{\percent} therefore indicates that every bin contains at least one data point.
For $v_\mathrm{int}$, a coarser partition of \num{20} bins was chosen rather than the \num{100} used for the other parameters: since the intruder speed varies only over the narrow interval, a \num{100}-bin partition would resolve this range too finely, leaving individual bins sparsely populated and in conflict with the expected-count prerequisite ($n\cdot p_i \geq 5$) established in \cref{sec:cov_metrics2}.
For both measures, the value column is accompanied by its categorical interpretation in accordance with \cref{tab:odd_thresholds}.
\begin{table}[htb]
\centering
\caption{Coverage report for the $ODD_{VCAS}$ with bin coverage, KL divergence and Cramér's $V$.}
\label{tab:vcas_coverage}
\begin{tabular}{lllcccccc}
\toprule
$ODD_{VCAS}$  & Nr.& Target & Range                  & Bin Cov.           & $D_\mathrm{KL}$ & $D_\mathrm{KL}$   & Cramér's $V$& Cramér's $V$  \\
          & Bins& Dist. & Util. (\unit{\percent}) & (\unit{\percent}) & Value    & Result &  Value  &  Result \\
\midrule
$h$       &     \num{100}     &  Normal & 92.96  & 89.65  & 1.0382 & Fail      & \num{0.1682} & Good \\
$\dot{h}_\mathrm{own}$& \num{100} &  Normal & 87.80  & 64.35  & 1.5507 & Fail      & \num{0.2357} & Moderate \\
$\tau$               & \num{100} &Uniform  & 374.85 & 100.00 & 0.0002 & Excellent & \num{0.0013} & Excellent \\
\bottomrule
\end{tabular}
\end{table}
\begin{table}[htb]
\centering
\caption{Coverage report for the $ODD_{HCAS}$ with bin coverage, KL divergence and Cramér's $V$.}
\label{tab:hcas_coverage}
\begin{tabular}{lllcccccc}
\toprule
$ODD_{HCAS}$  & Nr.& Target & Range                  & Bin Cov.           & $D_\mathrm{KL}$ & $D_\mathrm{KL}$   & Cramér's $V$& Cramér's $V$  \\
           & Bins & Dist. & Util. (\unit{\percent}) & (\unit{\percent}) & Value    & Result &  Value  &  Result \\
\midrule
$\rho$       & \num{100}    & Uniform & 173.95    & 100.00 & 0.0236 & Excellent   & 0.0212  & Excellent \\
$\theta$    &\num{100}  & Uniform & 100.00   & 99.41  & 1.1164 & Fail  &  0.2782  & Moderate\\

\(\psi\)    &\num{100}  & Uniform & 100.00   & 95.29  & 1.1599 & Fail  &  0.5563  & Fail\\

$v_\mathrm{int}$ &\num{20} & Uniform & 469.36    & 100.00  & 0.0444 & Excellent &  0.0202 & Excellent \\
$\tau$   & \num{100} & Uniform & 272.62   & 100.00 & 0.0019 & Excellent & 0.0060 &  Excellent \\
\bottomrule
\end{tabular}
\end{table}
For VCAS, $\tau$ shows a range utilization of \qty{374.85}{\percent}, substantially exceeding its baseline range and indicating that the collected data extend well beyond the originally specified ODD boundaries.
The parameters $h$ and $\dot{h}_\mathrm{own}$ reach \qty{92.96}{\percent} and \qty{87.80}{\percent}, reflecting adequate but incomplete coverage of their defined ranges.
A comparable pattern holds for HCAS: $\rho$, $v_\mathrm{int}$, and $\tau$ all exceed their baseline ranges (\qty{173.95}{\percent}, \qty{469.36}{\percent}, and \qty{272.62}{\percent}), whereas $\theta$ and $\psi$ each attain exactly \qty{100}{\percent}.
In terms of bin coverage, $\tau$ achieves \qty{100}{\percent} and $h$ reaches \qty{89.65}{\percent} for VCAS, while $\dot{h}_\mathrm{own}$ yields the lowest value at \qty{64.35}{\percent}, indicating that a notable portion of its input space remains sparsely populated, as also visible in \cref{fig:h_dot_down_plot}.
For HCAS, $\rho$, $v_\mathrm{int}$, and $\tau$ are fully covered, while $\theta$ reaches \qty{99.41}{\percent} and $\psi$ \qty{95.29}{\percent}.
The distributional measures provide a more differentiated picture than bin coverage alone.
The parameters with the broadest data support---$\tau$ for VCAS as well as $\rho$, $v_\mathrm{int}$, and $\tau$ for HCAS---are rated \textit{Excellent} by both $D_\mathrm{KL}$ and Cramér's $V$, with negligible divergences (below \num{0.05}) and correspondingly small association values, confirming that their empirical distributions closely match the assumed uniform baselines.
In contrast, $h$ and $\dot{h}_\mathrm{own}$ (VCAS) together with $\theta$ and $\psi$ (HCAS) obtain a \textit{Fail} KL rating (\num{1.0382}, \num{1.5507}, \num{1.1164}, and \num{1.1599}, respectively).
Under Cramér's $V$, $h$, $\dot{h}_\mathrm{own}$, and $\theta$ are assessed more favorably, with $h$ rated \textit{Good} (\num{0.1682}) and both $\dot{h}_\mathrm{own}$ (\num{0.2357}) and $\theta$ (\num{0.2782}) rated \textit{Moderate}.
The relative heading $\psi$, by contrast, is the only parameter rated \textit{Fail} by both measures (Cramér's $V = \num{0.5563}$), so that the two indicators concur on a substantial distributional mismatch.

\begin{figure}[htb]
\centering
\includegraphics[width=0.85\textwidth]{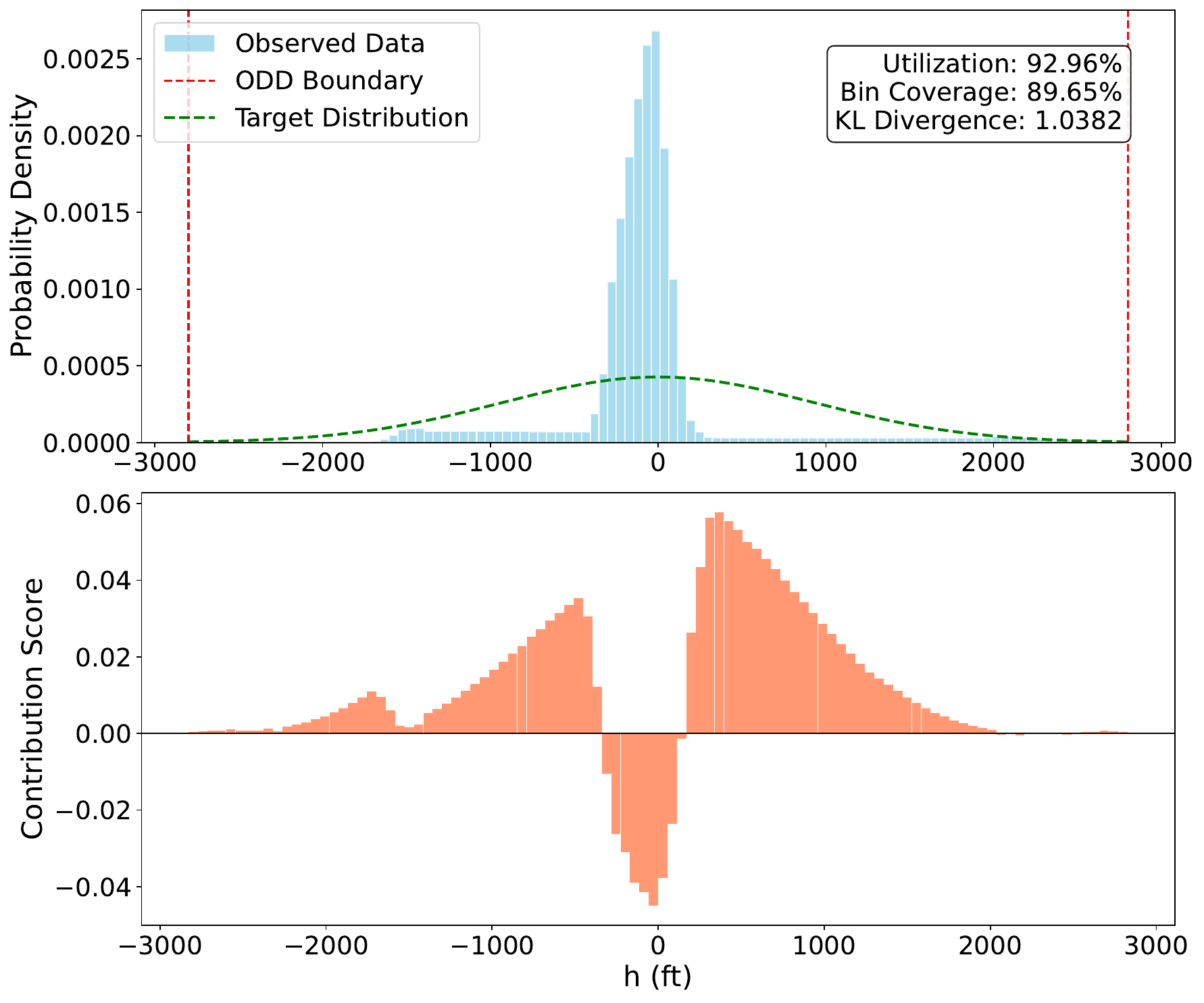}
\caption{Kullback--Leibler divergence of relative altitude $h$ from $ODD_{VCAS}$ against normal distribution.}
\label{fig:vcas_kl_div_h}
\end{figure}

\cref{fig:vcas_kl_div_h} illustrates the KL divergence assessment for the relative altitude $h$ of $ODD_\mathrm{VCAS}$ against the assumed normal target distribution.
The upper panel contrasts the observed data with the target density: the collected data are sharply concentrated around $h = \qty{0}{ft}$ and largely confined to the interval $[\qty{-500}{ft},~\qty{500}{ft}]$, whereas the target normal spreads its mass broadly across the full ODD range delimited by the boundaries at $\approx\qty{\pm2800}{ft}$.
The discrepancy is therefore one of concentration: the empirical distribution is far more peaked than the reference, despite both sharing the same central location.
The lower chart decomposes the total divergence into per-bin contributions and localizes the source of the mismatch.
The dominant contributions arise in the flanking regions, from a few hundred feet out to roughly \qty{2000}{ft} on either side of the center, where the target assigns appreciable probability mass that the observed data do not realize.
The densely populated central bins, in turn, form a narrow region of opposite-signed contribution. 
The accumulation of these flank terms yields the overall divergence of $D_\mathrm{KL} = \num{1.0382}$, rated with \textit{Fail}.
This \textit{Fail} rating coexists with a high range utilization of \qty{92.96}{\percent} and a bin coverage of \qty{89.65}{\percent}.
The data thus populate nearly the entire defined input space, yet in proportions that deviate strongly from the target.
This case demonstrates concretely that coverage and representativeness are distinct properties: complete coverage does not imply that the data follow the intended distribution---precisely the gap that the distributional measures are introduced to detect.

\begin{figure}[htb]
\centering
\includegraphics[width=0.88\textwidth]{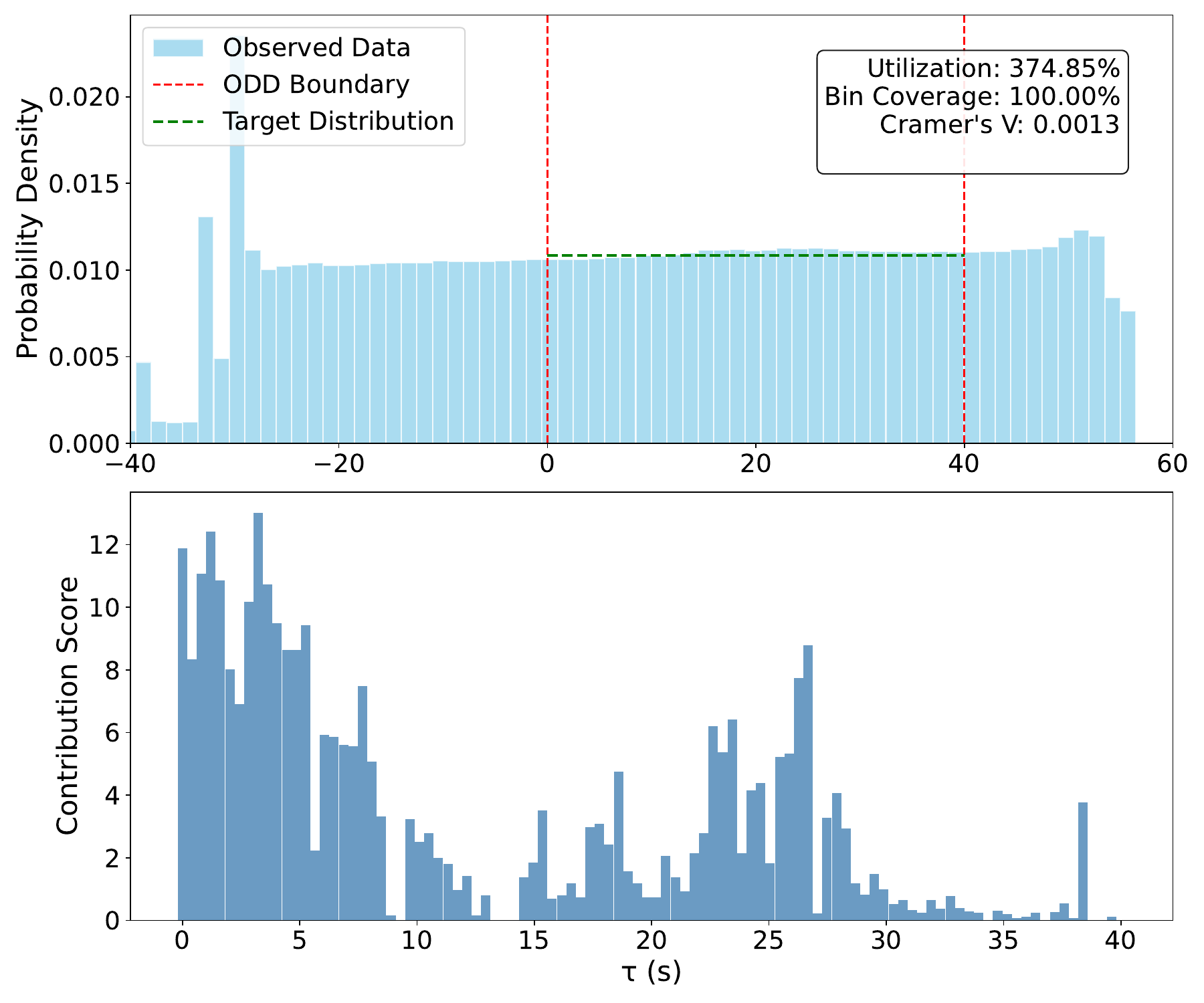}
\caption{Cramér's $V$ of time to CPA $\tau$ from $ODD_{VCAS}$ against uniform distribution.}
\label{fig:vcas_cramersV_h}
\end{figure}

\cref{fig:vcas_cramersV_h} shows the Cramér's $V$ assessment for the time to CPA $\tau$ of $ODD_\mathrm{VCAS}$ against the assumed uniform target distribution.
Within the ODD range delimited by the boundaries at $\tau = \qty{0}{s}$ and $\tau = \qty{40}{s}$, the observed density (upper panel) is essentially flat at $\approx\num{0.011}$ and closely tracks the uniform reference, in marked contrast to the strongly peaked altitude distribution of \cref{fig:vcas_kl_div_h}.
A substantial share of the collected data lies beyond the ODD boundaries---most visibly the negative-$\tau$ samples and the elevated bins near $\tau \approx \qty{-30}{s}$---consistent with the range utilization of \qty{374.85}{\percent}; the representativeness assessment is nonetheless evaluated only over the defined range, where the empirical and target distributions agree closely.
Beneath, the chart reports the per-bin contributions to the $\chi^2$ statistic underlying Cramér's $V$, which are non-negative by construction.
The largest contributions concentrate at short times to CPA (roughly $\tau \in [\qty{0}{s}, \qty{8}{s}]$), with a secondary cluster between $\qty{15}{s}$ and $\qty{30}{s}$, indicating that the mild residual deviations from perfect uniformity are localized rather than systematic.
Relative to the large in-range sample size, these deviations remain negligible, and the resulting Cramér's $V = \num{0.0013}$ is rated \textit{Excellent}.
Here, the strong coverage indicators are corroborated rather than contradicted by the distributional measure: the data fill the ODD range completely and also approximate the intended uniform distribution, so coverage and representativeness are jointly satisfied.
This contrasts with the altitude case in \cref{fig:vcas_kl_div_h}, where high coverage masked a substantial distributional mismatch.
Representativeness of a dataset with respect to an ODD parameter is demonstrated when the following two conditions hold, evaluated in order:
\begin{enumerate}
    \item \textbf{Bin-Coverage-Gate.} Every bin of the parameter's ODD partition contains at least one sample, i.e.\ the bin coverage equals \qty{100}{\percent}. The number of bins is fixed by the sensitivity analysis in \cref{fig:sensitivity_analysis} and is not adjusted to satisfy this criterion.
    \item \textbf{Divergence-Condition.} Provided condition~1 holds, the divergence of the empirical distribution from the target distribution remains within the tolerance assigned to the parameter's safety-relevance tier under both the KL divergence $D_\mathrm{KL}$ and Cramér's $V$, in accordance with the interpretation thresholds (e.g. \emph{Good} or better) listed in \cref{tab:odd_thresholds} and \cref{tab:cramersv}.
\end{enumerate}
A parameter satisfies the representativeness criterion only if both conditions are met.
Failure of condition~1 precludes the evaluation of condition~2, as the distributional measures are reliable only over a fully populated partition.
Finally, since representativeness is anchored to a chain of EASA's anticipated objectives, the corresponding mapping is summarized in \cref{tab:moc_mapping}~\cite{EUASA2024}.

\begin{table}[htb]
\centering
\caption{Mapping of the proposed coverage method to EASA's anticipated
objectives and means of compliance (Issue~02) for the data representativeness properties~\cite{EUASA2024}.}
\label{tab:moc_mapping}
\small
\begin{tabularx}{\textwidth}{@{}l X X@{}}
\toprule
EASA item & Data-quality requirement & Addressed by \\
\midrule
DA-04 & Capture of data quality requirements, incl.\ completeness and
representativeness, per ODD parameter, with documented origin and traceability &
Target-distribution definition and documentation (\cref{fig:target},
\cref{tab:target_distribution_rationale}) \\
\addlinespace
DM-07-2 & Verification that data are sampled per the prescribed distribution;
$\chi^2$/KS goodness-of-fit anticipated for low-dimensional operating parameters
& Representativeness assessment via $D_{\mathrm{KL}}$ and Cram\'er's $V$
(\cref{sec:results}) \\
\addlinespace
DM-08 & Data-verification step confirming ODD coverage with the necessary
completeness and representativeness & Two-stage coverage loop
(\cref{fig:MBSE_1}) \\
\bottomrule
\end{tabularx}
\end{table}

\section{Discussion}\label{sec:Disscusion}

This work introduces a novel, quantitative method for demonstrating ODD coverage in the learning assurance of safety-critical AI-based systems, developed in alignment with EASA's anticipated objectives~\cite{EUASA2024}.
ODD coverage rests on two complementary properties of the data set: representativeness, that the data follow the distribution prescribed by the ODD, required by Objective~DA-04 and Anticipated MOC~DM-07-2 and confirmed at the DM-08 verification step; and completeness, that the data span the ODD combinatorially, underlying the pairwise ODD-parameter coverage anticipated by Objective~LM-16.
The present work formalizes the representativeness property.
Consequently, the combinatorial completeness of the joint parameter space is not in the scope of this work and developed separately~\cite{Stefani2026}.
However, only the two pillars together operationalize the coupled representativeness and completeness expectations of EASA's learning-assurance building block.
Since demonstrating representativeness requires more than confirming that the data populate the ODD range, a parameter may span its entire range and fill nearly every bin while its empirical distribution still deviates substantially from the target.
Populating the input space, therefore, does not by itself establish representativeness; the proportions in which that space is sampled must also be assessed.
This motivates the two-stage structure of the proposed criterion, in which a population gate precedes the distributional evaluation.
The distributional assessment was initially based on the $\chi^2$ goodness-of-fit test, following its use for learning-assurance data analysis in the Collins/EASA study~\cite{ForMuLA}.
In our setting, this test proved unsuitable as a decision criterion.
Because the $\chi^2$ statistic scales with the sample size, the large data sets generated from the simulation drive its $p$-value toward zero for even negligible deviations, leading to rejection of the null hypothesis for every parameter---including $\tau$, whose empirical distribution is in excellent agreement with its uniform target under every effect-size measure.
A test that cannot pass even a well-matched parameter offers no discriminatory power and cannot serve as a representativeness gate.
We therefore adopted Cramér's $V$, which normalizes the same $\chi^2$ statistic by the sample size and the degrees of freedom to yield a bounded effect size.
We retain the KL divergence alongside Cramér's $V$ because the two measures capture complementary aspects of distributional agreement.
The KL divergence is an information-theoretic measure that is particularly sensitive to regions where the target assigns probability mass the data do not realize, making it responsive to shape and tail mismatches, whereas Cramér's $V$ summarizes the overall magnitude of deviation on a bounded, sample-size-independent scale. Requiring both to fall within tolerance guards against failure modes that either measure alone could miss.
Notably, the two measures concur for every parameter that satisfies the bin-coverage gate. Among the parameters with unpopulated bins, $\psi$ is rated \textit{Fail} by both measures, whereas for $h$, $\dot{h}_\mathrm{own}$, and $\theta$, the KL divergence is inflated by the empty-bin contributions to which it is inherently sensitive, so that it fails these parameters while the bounded Cramér's $V$ rates them only \textit{Good} to \textit{Moderate}.
This pattern both supports the joint use of the two measures and reinforces the condition ordering: each statistic is trusted only within the regime in which it is reliable---$\chi^2$ away from very large samples, the KL divergence away from sparse partitions.
Several limitations remain within the representativeness pillar addressed here.
The bin-coverage-gate in its current form requires only that each bin be populated,
which guarantees that the ODD range is sampled but not that the per-bin frequencies are estimated with sufficient statistical reliability.
Consequently, determining the adequate number of bins is of great importance to account for the required level of granularity.
The tolerance bands for Cramér's $V$ are currently mapped to qualitative interpretation ranges and are not yet anchored to the safety-relevance tier of each parameter, which is required before the criterion can support a formal certification argument.
However, the assessment is only as meaningful as the assumed target distributions: the \textit{Fail} ratings obtained for $h$ and $\dot{h}_\mathrm{own}$ partly reflect the broad normal reference imposed on parameters whose realistic encounters concentrate near small relative altitudes, and replacing these assumed references with empirical distributions derived from operational surveillance data would  strengthen the representativeness claim.
Finally, this work provides a generic and reusable coverage assessment loop that is aligned with EASA's guidelines for AI-based systems in aviation, in particular with respect to the data representativeness objectives outlined in the learning assurance building block.
The results confirm the practical usability of the proposed method and highlight that no single metric is sufficient for a comprehensive ODD coverage assessment.
A combination of bin coverage and distributional similarity measures, such as KL divergence and Cramér's $V$, provides a more complete and reliable picture of data set quality for safety-critical AI/ML applications.

\section{Conclusion and Outlook}\label{sec:Conclusion}

This work addressed the assessment of data representativeness for the ODD of safety-critical AI-based systems---the property required by Objective~DA-04 and Anticipated MOC~DM-07-2 and verified, together with completeness, at the DM-08 data-verification step.
We proposed a two-stage representativeness criterion that couples a data population gate over the ODD partition with a divergence evaluation based on the KL divergence and Cramér's $V$, and demonstrated it on the neural-network-based HCAS and VCAS.
The case study supports three conclusions.
First, the $\chi^2$ goodness-of-fit test loses its usability on the large simulation-derived data sets considered here and cannot serve as a decision criterion; sample-size-robust effect-size measures are required instead.
Second, no single metric is sufficient: bin coverage establishes that the ODD range is populated, whereas the KL divergence and Cramér's $V$ capture complementary aspects of distributional agreement, and their joint use yields a more reliable picture of data set quality than any measure alone.
Third, the resulting assessment loop is generic and reusable, and maps directly onto the data-representativeness expectations of the EASA learning-assurance building block.

Several directions extend this work.
The most immediate is the formalization of the second pillar, completeness, which evaluates the combinatorial coverage of the joint ODD space, and which the present criterion does not yet address.
The bin-coverage-gate can be strengthened from a minimal occupancy requirement to a statistically grounded one: a per-bin sample-size requirement calibrated by a power analysis is a direction for future work, ensuring not only that each bin is populated but that the per-bin frequencies are estimated reliably enough for the distributional measures to be trustworthy.
A complementary refinement concerns the acceptance thresholds, where the qualitative interpretation ranges currently used for the KL divergence and Cramér's $V$ would be replaced by application-specific acceptance ranges that map value interpretations to the safety criticality of each parameter, holding safety-relevant parameters to tighter tolerances; such a tier-anchored mapping is a precondition for supporting a formal certification argument.
Finally, the assumed target distributions can be replaced by empirical references derived from operational data, and metrics better suited to ordered continuous variables can be investigated as complements to the present measures.

\section{Contact Author Email Address}
\underline{\href{mailto:thomas.stefani@dlr.de}{thomas.stefani@dlr.de}}

\section{Copyright Statement}
\begin{small}
The authors confirm that they, and/or their company or organization, hold copyright on all of the original material included in this paper. The authors also confirm that they have obtained permission, from the copyright holder of any third party material included in this paper, to publish it as part of their paper. The authors confirm that they give permission, or have obtained permission from the copyright holder of this paper, for the publication and distribution of this paper as part of the ICAS proceedings or as individual off-prints from the proceedings.
\end{small}

\biblio{literature-bibtex}
\label{bibliography}

\end{document}